\documentclass[10pt,twocolumn,letterpaper]{article}

\usepackage{iccv}
\usepackage{times}
\usepackage{epsfig}
\usepackage{graphicx}
\usepackage{amsmath}
\usepackage{amssymb}
\usepackage{graphicx}
\usepackage{wasysym}
\usepackage{algorithm}
\usepackage{algpseudocode}

\usepackage[breaklinks=true,bookmarks=false]{hyperref}

\iccvfinalcopy 

\def\iccvPaperID{7532} 

\ificcvfinal\fi

\begin{document}

\title{Pseudo-mask Matters in Weakly-supervised Semantic Segmentation}

\author{Yi Li$^{1}$\and Zhanghui Kuang$^{1}$\and Liyang Liu$^{2}$\and Yimin Chen$^{1}$\and Wayne Zhang$^{1,3,4}$ \and
SenseTime$^{1}$\and Tsinghua University$^{2}$\and Qing Yuan Research Institute, Shanghai Jiao Tong University$^{3}$\and Shanghai AI Laboratory$^{4}$\and
{\tt\small \{liyi,kuangzhanghui,chenyimin,wayne.zhang\}@sensetime.com}, {\tt\small liu-ly14@mails.tsinghua.edu.cn}\and
}

\maketitle
\ificcvfinal\thispagestyle{empty}\fi

\begin{abstract}
Most weakly supervised semantic segmentation (WSSS) methods follow
the pipeline that generates pseudo-masks initially and trains the
segmentation model with the pseudo-masks in fully supervised manner
after. However, we find some matters related to the pseudo-masks, including high quality pseudo-masks generation from class activation maps (CAMs), and training with noisy pseudo-mask supervision. For these matters, we propose the following designs to push the performance to new state-of-art: (\romannumeral1) Coefficient of Variation Smoothing to smooth the CAMs adaptively; (\romannumeral2) Proportional Pseudo-mask Generation to project the expanded CAMs to pseudo-mask based on a new metric indicating the importance of each class on each location, instead of the scores trained from binary classifiers.  (\romannumeral3) Pretended Under-Fitting strategy
to suppress the influence of noise in pseudo-mask; (\romannumeral4) Cyclic Pseudo-mask to boost the pseudo-masks during training of fully supervised semantic segmentation (FSSS). Experiments based on our methods achieve new state-of-art results on two changeling weakly supervised semantic segmentation datasets, pushing the mIoU to 70.0\% and 40.2\% on PAS-CAL VOC 2012 and MS COCO 2014 respectively. Codes including segmentation framework are released at \href{https://github.com/Eli-YiLi/PMM}{https://github.com/Eli-YiLi/PMM}
\end{abstract}

\section{Introduction}


Semantic segmentation is a fundamental computer vision problem and
requires time-consuming pixel-level manual annotations. To reduce
the annotation burden, weakly-supervised semantic segmentation approaches
have been proposed using scribble annotations \cite{lin2016scribblesup,Tang_2018_CVPR},
bounding boxes \cite{Xu_2015_CVPR,dai2015boxsup,Khoreva_2017_CVPR}
, points \cite{bearman2016s} or image-level labels \cite{kolesnikov2016seed,pathak2015constrained,pinheiro2015image,ahn2018learning,Wang_2020_CVPR}.
In this paper, we focus on weakly-supervised semantic segmentation
with image-level labels due to its easily available annotations.


Almost all the latest WSSS algorithms require pseudo-mask derived
from CAM to train the FSSS model. Instead of the pseudo-mask, previous
works mainly focus on the generation of CAMs, or the post-process of
them. 
We observe that there are some matters about pseudo-mask that
are not handled appropriately as follows: (\romannumeral1) argmax operation on the CAMs along the channel dimension projects multi-label class activation
maps (CAMs) to single-label pseudo-masks,
but the image-level classification for generating CAMs does not consider the conflicts of predictions on target locations;
(\romannumeral2) CAMs generated by the classification model tend
to focus on the most discriminative part and result in partial activations;
(\romannumeral3) the noise in pseudo-masks is inevitable and impedes
the training of fully-supervised semantic segmentation (FSSS). (\romannumeral4)
the predictions of FSSS are usually more accurate than supervisory signals (pseudo-masks).


In this paper, we propose a series of strategies to boost the efficiency
of pseudo-masks in aspects of both generation and utilization. Specifically,
in the pseudo-mask generation step, we firstly compute the Coefficient
of Variation ($c_{v}$) for each channel of CAMs, and then refine CAMs
via exponential functions with $c_{v}$ as the control coefficient.
This operation smooths the CAMs and could alleviate the partial response
problem introduced by the classification pipeline. Instead of projecting
the three-dimensional CAM after dense-CRF \cite{NIPS2011_beda24c1}
directly to two-dimensional pseudo-mask with the argmax operation
on scores as in previous studies, we equip each pixel a scalar which
is computed as the proportion between the pixel's attention and the
attention sum of the channels over the whole image. Intuitively the scalar represents the importance of the corresponding pixel
based on which the final pseudo-masks are generated. In the FSSS training
step, we propose a Pretended Under-fitting Strategy which suppresses the
losses of the noise labels in the pseudo-masks. In addition, the model
is evaluated on validation dataset and we update the masks cyclically
in condition that prediction from model is better than the pseudo-masks,
rather than use the fixed pseudo-masks generated in the first step.


Applying our methods to a baseline algorithm called SEAM \cite{Wang_2020_CVPR},
we achieve new state-of-the-art results on two challenging weakly-supervised
semantic segmentation benchmarks. In particular, our approach reach
the mIoU of 70.0\% and 40.2\% on In PASCAL VOC 2012 \cite{everingham2010pascal}
and MS COCO 2014 \cite{lin2014microsoft} validation sets respectively.

The contributions of this paper are three-fold: 
\begin{itemize}
\item We generate high-quality pseudo-masks by the proposed Proportional
Pseudo-mask Generation with Coefficient of Variation Smoothing. The
Coefficient of Variation Smoothing expands the activation area of
CAMs to overcome the partial response problem based on the CAM's coefficient
of variation, and the Proportional Pseudo-mask Generation computes
the importance of each location for each class independently, based
on which the final pseudo-masks are generated. 
\item We realize the effective utilization of pseudo-masks via reducing
the influence of noise by our Pretended Under-fitting Strategy, and narrow the gap between ground-truths and pseudo-masks via Cyclic Pseudo-masks.
The pixel-wise losses are reweighted to suppress the noises in the
Pretended Under-fitting Strategy and pseudo-masks are refined in a
iterative manner. 
\item We conduct extensive experiments to validate the effectiveness of
our proposed approach (Pseudo-mask Matters, PMM), and demonstrate
that our approach achieves new state-of-the-art results on two challenging
weakly-supervised semantic segmentation datasets. 
\end{itemize}

\section{Related Work}

\textbf{Weakly Supervised Semantic Segmentation:} What WSSS does is
to simplify the supervision with less accuracy loss. The annotation
cost weakens from mask \cite{chen2017deeplab,zhao2017pyramid} (fully
supervision) to scribble \cite{lin2016scribblesup,Tang_2018_CVPR},
bounding box \cite{Xu_2015_CVPR,dai2015boxsup,Khoreva_2017_CVPR},
points \cite{bearman2016s} and image label \cite{kolesnikov2016seed,pathak2015constrained,pinheiro2015image,ahn2018learning,Wang_2020_CVPR}
gradually. Till image-level label, there is only category information
without spatial supervision. Thanks to the translation invariance
of CNNs, the classification contributed pixels keep high response
in the feature map. After combining the weights of classifier with
the feature map, we get the CAM \cite{zhou2016learning} as the
initial semantic mask. Later works mostly focus on expanding the seed
areas. The methods include adding CRF and dilation
\cite{wei2018revisiting}, designing new losses \cite{kolesnikov2016seed},
erasing high response area \cite{hou2018self}, keeping scale consistency
via siamese network and correlation module \cite{Wang_2020_CVPR},
clustering sub-categories for classification \cite{chang2020weakly}.
Besides CAM based methods, weakly supervised object detection method
is also used with proposal models \cite{liu2020leveraging}.


For the overall pipeline of WSSS, firstly a binary classification model is trained to obtain the CAMs and several techniques have been proposed to improve its quality. Secondly post-processing has been applied on the CAMs to generate pseudo-masks. Finally, semantic segmentation model with pseudo-masks supervised is trained in fully supervised manner.


\textbf{Pseudo-mask Generation:} Pseudo-mask generation is to project
three-dimensional CAM to two-dimensional pseudo-mask with some post-process algorithms.
Following the manner of single label classification, the pixel-level labels are
identified by argmax operation on the CAMs along the channel dimension after post-process, although
the model to predict the CAMs is trained by binary classifiers which exclude the label competition.

Conditional Random Fields (CRF), a sort of statistical modeling method for structured prediction
with considering neighboring samples, has been widely used as the post processing tool for segmentation. Some variations of CRF have been proposed. Dense-CRF \cite{NIPS2011_beda24c1} applies the appearance
kernel to link nearby pixels with similar color, and smoothness kernel
to removes small isolated regions. Deeplabv1 \cite{chen2015semantic}
is an early work that introduces fully connected CRF in to segmentation
task as post process. Then, CRFasRNN \cite{zheng2015conditional}
combines the CNN and CRF end-to-end. Besides, DPN \cite{liu2015semantic}
uses the MRF which is similar to CRF. Further more, G-CRF \cite{chandra2016fast} introduced CNN for potential learning to improve the performance.

Several post-process methods based on deep learning have also been proposed.
\cite{ahn2018learning,8953768} learn the contour of objects
via an affinity network in inference stage. The training
masks are synthesised from two CRF results with different background
intensities. Besides the unsupervised contour learning method, other
works introduce saliency detection models trained from extra dataset
to refine CAMs \cite{yao2020saliency,zeng2019joint,fan2020learning}.

\section{Methodology}

\begin{figure*}
\begin{centering}
\includegraphics[scale=0.52]{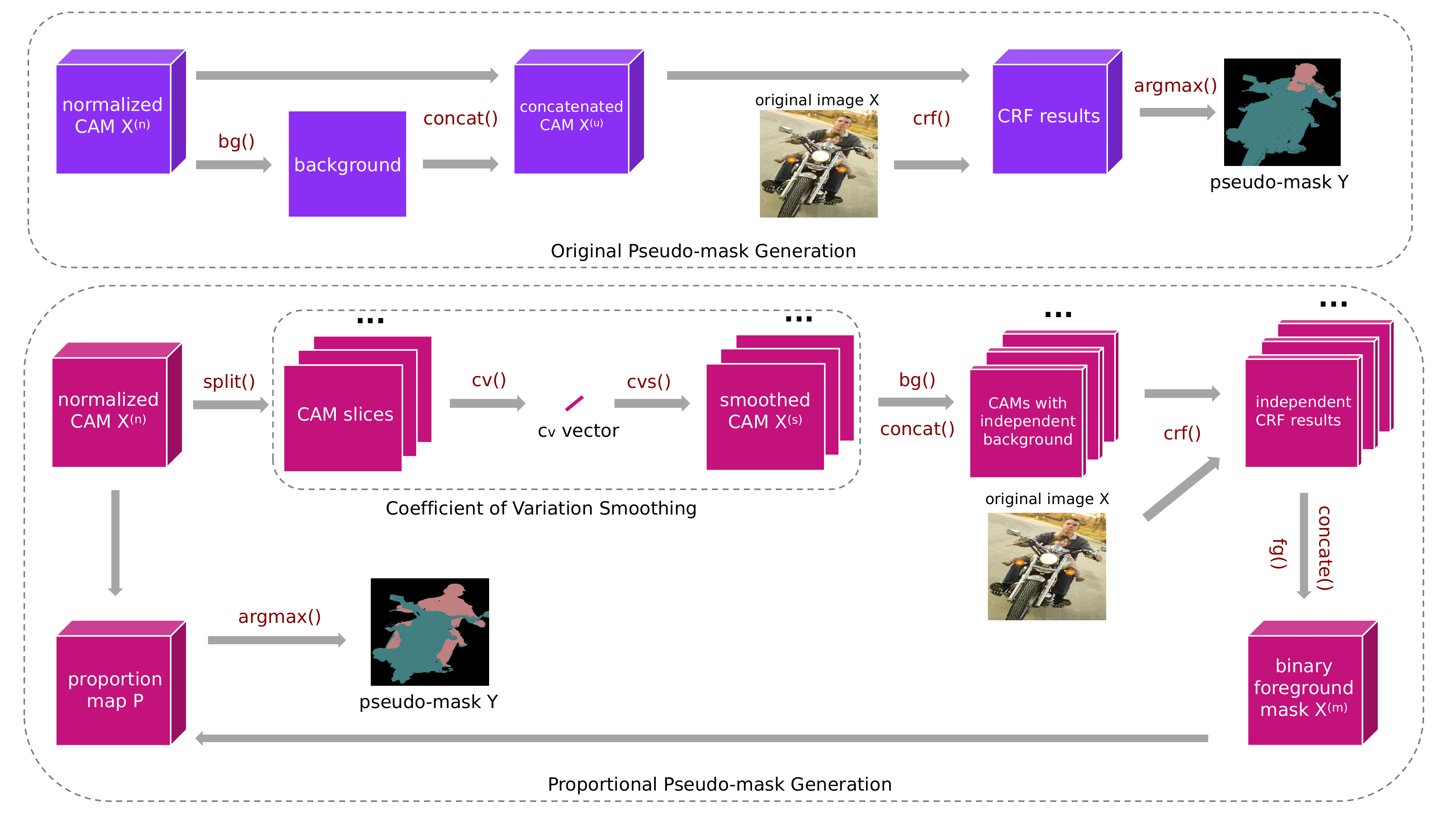}
\par\end{centering}
\caption{\label{fig:generation}Illustration of pseudo-mask generation. Top:
traditional pipeline for pseudo-mask generation. Bottom: Proportional Pseudo-mask
Generation with Coefficient of Variation Smoothing. $cv$ counts the
coefficient of variation for each channel and $cvs$ smooths the CAM.
$bg$ generates the background with exponential function and $fg$
returns the binary foreground. $p$ counts the proportion of a pixel
to the entire category.}
\end{figure*}

In this section, We firstly elaborate the traditional pipeline generating pseudo-mask
which projects the class activation map $\boldsymbol{X}\in\mathbb{R}^{C\times H\times W}$
to pseudo-mask $\boldsymbol{Y}\in\mathbb{R}^{H\times W}$ with CRF
function $crf$ and image $\boldsymbol{I}$. Then we introduce our pseudo-mask generation method, Proportional
Pseudo-mask Generation (PPMG) with Coefficient of Variation Smoothing
(CVS) as Fig. \ref{fig:generation}.
After that, the noise suppressing module, Pretended Under-fitting
Strategy (PUS) for the FSSS is described. Finally, the overall pipeline with Cyclic
Pseudo-masks (CPM) involved is presented.
We name our pipeline PMM from the abbreviation of Pseudo-mask Matters.

\subsection{Pseudo-mask Generation with CRF}

The CRF algorithm $crf$ here is dense-CRF \cite{NIPS2011_beda24c1}.
The CAMs are processed with operations, including 
normalization $norm$, exponentially background generation
$bg$ and $argmax$, and the results are set as the input to CRF.

Specifically, Min-Max Normalization is 
applied on $\boldsymbol{X}_{c,:,:}$
and formulated as: 
\begin{equation}
norm(\boldsymbol{X}_{c,h,w})=\frac{\boldsymbol{X}_{c,h,w}-min(\boldsymbol{X}_{c,:,:})}{max(\boldsymbol{X}_{c,:,:})-min(\boldsymbol{X}_{c,:,:})},\forall c,h,w\label{eq:norm}
\end{equation}
where $h$ and $w$ are the coordinates on the CAMs, and $c$ represents the channel index.

After normalization for each pixel, we have $\boldsymbol{X}^{(n)}$.
Then the normalized pixel $\boldsymbol{X}_{c,h,w}^{(n)}$ on background
matrix is constructed exponentially with power $\alpha$ from the
normalized foreground pixels $\boldsymbol{X}_{:,h,w}^{(n)}$ in location
$h$ and $w$: 
\begin{equation}
bg(\boldsymbol{X}_{:,h,w}^{(n)})=(1-max(\boldsymbol{X}_{:,h,w}^{(n)}))^{\alpha},\forall h,w
\end{equation}


Then concatenate the background and foreground to form the input $\boldsymbol{X}^{(u)}$
of unary potential function in CRF: 
\begin{equation}
\boldsymbol{X}^{(u)}=concat(bg(\boldsymbol{X}^{(n)}),\boldsymbol{X}^{(n)})
\end{equation}

Finally, the pseudo-mask is identified via argmax operation in category
channel after CRF: 
\begin{equation}
\boldsymbol{Y}=argmax(crf(\boldsymbol{I},\boldsymbol{X}^{(u)})),\label{argmax}
\end{equation}

\subsection{Coefficient of Variation Smoothing}

The motivation of Coefficient of Variation Smoothing (CVS) is to smooth
the class activation map based on the variation of confidence in spatial domain.
We believe different images and classes require varied smoothing intensity
depending on its distribution of confidences. To measure it, we introduce the
Coefficient of Variation ($c_{v}$) as the metric for the foreground
pixels $\boldsymbol{X}_{c,:,:}^{(f)}$ whose scores are higher than
threshold $t$ in normalized metric $\boldsymbol{X}_{c,:,:}^{(n)}$
at channel $c$. We define the $cv$ function in Eq.\eqref{CoV} where
$\mathbb{D}$ counts the deviation $\sigma^{2}$ and $\mathbb{E}$
counts the mean $\mu$. 
\begin{equation}
cv(\boldsymbol{X}_{c,:,:}^{(n)})=\frac{\sqrt{\mathbb{D}(\boldsymbol{X}_{c,:,:}^{(f)})}}{\mathbb{E}(\boldsymbol{X}_{c,:,:}^{(f)})},\label{CoV}
\end{equation}

Then the $c_{v}$ is used as exponential function power to each pixel.
As $\boldsymbol{X}^{(n)}\in[0, 1]$, lower exponential power under 1 leads smaller differences between the
foreground pixels and smooths the CAM. Here, We define the $cvs$ for
each pixel with scale factor $s$ as: 
\begin{equation}
cvs(\boldsymbol{X}_{c,h,w}^{(n)},(c_{v})_{c})=(X_{c,h,w}^{(n)})^{(1-s\times(c_{v})_{c})},\forall c,h,w
\end{equation}

The CVS is applied after the normalization operation Eq.(\ref{eq:norm}), and
expand the activation area of target objects. Experimentally,
CVS works more efficiently with stronger augmentations involved, which also requires more training iterations.

\subsection{Proportional Pseudo-mask Generation}

One important concern in WSSS is that the CAMs are obtained from
binary classifiers. Follow the independent manner of binary cross
entropy loss, we generate class-specific background for the smoothed
CAM $\boldsymbol{X}_{c,:,:}^{(s)}$ by $bg$ and apply $crf$ for
it, then we return the binary foreground via $fg$, thus we have $\boldsymbol{X}_{c,:,:}^{(m)}$
as:

\begin{equation}
\boldsymbol{X}_{c,:,:}^{(m)}=fg(crf(\boldsymbol{I},bg(\boldsymbol{X}_{c,:,:}^{(s)}))),\forall c
\end{equation}

In this function, $fg$ sets the foreground pixels whose scores are
upper than $t$ to 1, and other pixels to 0. So $\boldsymbol{X}_{c,:,:}^{(m)}$
is a binary class mask. Then we concatenate the masks to $\boldsymbol{X}^{(m)}$.

However, more than one labels may be assigned to some pixels as activation areas of different classes could be overlapped.
We do not assign the label to each pixel by activation scores on CAM, because in
binary cross entropy, the loss only requires the network to distinguish
positive sample and negatives for each category independently, and thus highlights the foreground consequently.
But scores of different categories on foreground areas are not competitive during the training of binary classification. So rather than taking the index
of the highest score as pseudo label, we compute a metric indicating the importance of each class on each pixel, based on which the pixel is assigned with more important label. Specifically, the CRF map is converted to a binary mask with the a thresholding operation and the computation of metric mention before could not be affected by the CRF score. The proportion function
$p$ is defined as:

\begin{equation}
p(\boldsymbol{X}_{c,h,w}^{(n)})=\frac{\boldsymbol{X}_{c,h,w}^{(n)}}{sum(\boldsymbol{X}_{(c,:,:)}^{(n)}\cdot\boldsymbol{X}_{c,:,:}^{(m)})},\forall c,h,w\label{eq:proportion}
\end{equation}

In Eq.(\ref{eq:proportion}) $\boldsymbol{X}_{c,:,:}^{(m)}$ is the
foreground binary mask in channel $c$. 
Then, argmax is operated on the element-wise multiplication of mask and proportion map along channel dimension to generate pseudo-mask and formulated as:

\begin{equation}
\boldsymbol{Y}=argmax(\boldsymbol{X}^{(m)}\cdot\boldsymbol{P})\label{eq:allocate}
\end{equation}

Thus, each pixel is equipped with a single label after the processing above and could be serve as supervision for semantic segmentation training.


The pseudo implementation of Proportional Pseudo-mask Generation is described in  Alg.(\ref{alg:algorithm})
:

\begin{algorithm}
\caption{Proportional Pseudo-mask Generation\label{alg:algorithm}}

\textbf{Input}: image $\boldsymbol{I}$ and CAM $\boldsymbol{X}\in\mathbb{R}^{C\times H\times W}$

\textbf{Output}: pseudo-mask $\boldsymbol{Y}\in\mathbb{R}^{H\times W}$

$\,\,$1: normalize the CAM: $\boldsymbol{X}^{(n)}=norm(\boldsymbol{X})$

$\,\,$2: count $c_{v}$ for each class: $c_{v}=cv(\boldsymbol{X}^{(n)})$

$\,\,$3: smooth the CAM: $\boldsymbol{X}^{(s)}=cvs(\boldsymbol{\boldsymbol{X}}^{(n)},c_{v})$

$\,\,$4: compute binary mask with $crf$:

$\,\,\,\,\,\,\,$$\boldsymbol{X}_{c,:,:}^{(m)}=fg(crf(\boldsymbol{I},bg(\boldsymbol{X}_{c,:,:}^{(s)}))),\forall c$

$\,\,$5: count the proportion map: $\boldsymbol{P}=p(\boldsymbol{X}^{(n)})$

$\,\,$6: generare pseudo-mask: $\boldsymbol{Y}=argmax(\boldsymbol{X}^{(m)}\cdot\boldsymbol{P})$ 
\end{algorithm}

\subsection{Pretended Under-fitting Strategy}


Compared with the manual annotations, pseudo-masks that serves as supervision signal to train the semantic segmentation are noisy. Previous studies focus on the generation of high quality pseudo-masks to reduce the noise and rare of them attempt to suppress the noise during the model training.  
Our approach is to reweight the losses of potential noise
pixels in the optimization of FSSS. For this goal, we propose the
Pretended Under-fitting Strategy as Eq.(\ref{eq:ifl}):

\begin{equation}
\ell(\boldsymbol{L})=\begin{cases}
mean(\boldsymbol{L}) & mean(\boldsymbol{L})>=\beta\\
mean(pus(\boldsymbol{L})) & mean(\boldsymbol{L})<\beta
\end{cases}\label{eq:ifl}
\end{equation}


$\boldsymbol{L\in}\mathbb{R}^{H\times C}$ is
the loss map for pixels in the image from cross entropy loss, and
$\ell$ means the loss of Pretended Under-fitting Strategy for the
loss map $\boldsymbol{L}$. Funtion $pus()$ is the operation of Pretended
Under-fitting Strategy if the mean value of $\boldsymbol{L}$ below
warm up threshold $\beta$.

Three operations are implemented as followed:

\begin{equation}
pus_{clamp}(\boldsymbol{L})=\begin{cases}
\boldsymbol{L}_{h,w} & \boldsymbol{L}_{h,w}<\kappa\\
\boldsymbol{L}_{h,w}\cdot\frac{\kappa}{\boldsymbol{L}_{h,w}} & \boldsymbol{L}_{h,w}>=\kappa
\end{cases}\forall h,w\label{eq:clamp}
\end{equation}

\begin{equation}
pus_{pow}(\boldsymbol{L})=\boldsymbol{L}^{\kappa}\label{eq:pow}
\end{equation}

\begin{equation}
pus_{ignore}(\boldsymbol{L})=\begin{cases}
\boldsymbol{L}_{h,w} & \boldsymbol{L}_{h,w}<\kappa\\
0 & \boldsymbol{L}_{h,w}>=\kappa
\end{cases}\forall h,w\label{eq:ignore}
\end{equation}

$pus_{clamp}$ sets a maximum of $\boldsymbol{L}$ to a hyper parameter
$\kappa$ while $pus_{ignore}$ drops these pixels. $pus_{pow}$ carries
out a scaling strategy by exponential function. We later evaluate
these operations in Tab. \ref{tab:ifl}.

\subsection{Cyclic Pseudo-mask and Overall Pipeline}
\label{subsec::method_cycle}


Obviously there is a large gap between pseudo-mask and
real ground-truth. Previous studies have shown that CNN model is robust to noise in some degree. So a simple but effective method to narrow the
gap is to update the pseudo-mask cyclically. We replace the pseudo-mask on training dataset as the predictions from the trained model on it. We
call the new mask as Cyclic Pseudo-mask. This operation is validated in next section.


The overall pipeline is consisted of (\romannumeral1) classification for CAMs, (\romannumeral2) pseudo-mask
generation and (\romannumeral3) training of FSSS. In the first step, multi-crop test is used to generate the CAMs instead of multi-scale test. Specifically, Multi-crop firstly resizes
the image to different resolutions and crop the resized images with
fixed crop size and stride, then compute the average of crop results.
Meanwhile, we propose a multi-crop training technique which transforms
the image in the similar way to the test. Note that, we need a base mask to
update the ground truth, as multi-crop may crops the background
if the image is resized too large, causing the original labels invalid.
In our study, the CAMs from SEAM \cite{Wang_2020_CVPR} are used
as the base mask. The scale-friendly CNN structure, ScaleNet \cite{li2019data}, is applied to train the classification model and multi-crop operation which loads the rough mask and update
the image level label is implemented online.


\section{Experiments}

\subsection{Implementation Setup}

\textbf{Datasets}: We evaluate our approach on PASCAL VOC 2012 dataset
and MS-COCO 2014 dataset. All the mask annotations are converted to
image level multi-label ground-truths. VOC12 contains 20 foreground
objects and one background. The dataset is divided into train set
(1464 images), validation set (1449 images) and test set (1456 images).
In general, additional annotations from SBD \cite{hariharan2011semantic}
are used to augment training set to 10582 images. COCO14 dataset ranges
from 1 to 90, among them 80 categories are valid foreground. The train
set contains 82081 images and the number of validation set is 40137.
We evaluate the experiments on the validation sets by Mean intersection
over union (mIoU).

\textbf{Implementation details: }Our baseline of classification framework
is SEAM. We replace the backbone from Wide ResNet38 \cite{wu2019wider}
in SEAM to ScaleNet101 \cite{li2019data} to boost the multi-scale
feature which is similar to multiple resolution training in SEAM.
The output stride is 8 without extra dilations since the receptive
fields of ScaleNet is massive. Feature maps from stage 3 and stage
4 are projected to 64 and 128 channels respectively by 1$\times$1
convolution layers for the PCM module in SEAM.

The resize scales of multi-crop start from 0.75 to 3 (step 0.25).
The resized images are cropped in size 448$\times$448 with crop stride
300. In train phase, if crop area covers over 10\% of foreground class
$c$ or over 10\% area of crop is class $c$, we tag the ground-truth
to positive in channel $c$. The training images is randomly selected
from the multi-crop proposals. In test phase, the resize scales, crop
size and stride are the same to train phase. We get the base mask from
original SEAM with $\alpha$ in CRF setting to be 24 without Random Walk. 
SEAM is trained as original settings.
Note that multi-crop training requires 20 epochs to make sure that each cropped patch of images are involved. The initial learning rate:wq is set to 0.02
with batch size 16. The CAM background exponent $\alpha$ and the scale
factor $s$ of $cvh$ function are obtained from grid search (eg.
11 and 0.3), with foreground threshold $t$ at 0.05.

For the fully supervised segmentation, we do not reproduce the performance
of deeplab-v2 \cite{chen2017deeplab} as described in \cite{Wang_2020_CVPR}, so we use the PSPnet \cite{zhao2017pyramid} to achieve the mIoU in the paper with codebase MMSegmentation
\cite{mmseg2020}. We follow the settings in MMSegmentation for VOC, and we add dense-CRF to it. For Pretended Under-fitting Strategy, set the warm up threshold $\beta$ and hyper parameter $\kappa$
in PUS to 0.5 for VOC, and 0.8 for COCO. The training batch size is 16 on 8 gpus at learning
rate 0.005 for 20000 iterations in ploy policy. 
On VOC12 dataset, the pseudo-mask is updated once according to the method described in \ref{subsec::method_cycle} and we do only apply the pretended under-fitting strategy in the 1st round training.
On COCO14 dataset, the pseudo mask is not updated as we observe that the performance on validation dataset is the best after the 1st round training.
We set the class number to 91 (one
background and 10 invalid), and the invalid classes are ignored in
test. Besides that, we use 32 gpus, batch size 64, learning rate 0.02
and iteration 40000 for the training of COCO14.
All of the backbones in FSSS are initialized with the pretrained model from ImageNet\cite{krizhevsky2012imagenet}.

\subsection{Ablation Studies}

We make ablation studies for CAM generation, pseudo-mask generation
and segmentation in different settings. All the results are evaluated
on validation set of VOC12 for the consistency of comparison.

\textbf{Improvements of CAM}: In this paper, we push out a strong
CAM baseline based on multi-crop and multi-scale network. In Tab.
\ref{cam}, the top part evaluates the effectiveness of multi-crop
strategy. Notable, there is 3.49 improvement when multi-crop is applied
on both train and test phase, compared to 1.21 in test phase only. Then
we compare multi-scale backbones in the middle part, and we find the
ScaleNet101 performs significantly better than Res2Net101 and wide
ResNet38, which suggests that apply multi-scale operations in backbone
benefits CAMs a lot. We add multi-crop to train phase in last line
for ScaleNet101, and the final result is 58.21. 

\begin{table}
\begin{centering}
\begin{tabular}{ccc}
\hline 
Setting  & Backbone  & \multicolumn{1}{c}{VOC12 val}\tabularnewline
\hline 
\hline 
baseline  & ResNet38  & 52.72\tabularnewline
multi-crop test  & \multicolumn{1}{c}{ResNet38} & 53.93\tabularnewline
multi-crop test \& train  & ResNet38  & 56.21\tabularnewline
\hline 
multi-crop test  & Res2Net101  & 54.90\tabularnewline
multi-crop test  & ScaleNet101  & 57.81\tabularnewline
\hline 
multi-crop test \& train  & ScaleNet101  & \textbf{58.21}\tabularnewline
\hline 
\end{tabular}
\par\end{centering}
\caption{\label{cam}CAM mIoU of different settings.}
\end{table}

\begin{figure}
\begin{centering}
\includegraphics[width=2.8cm,height=2cm]{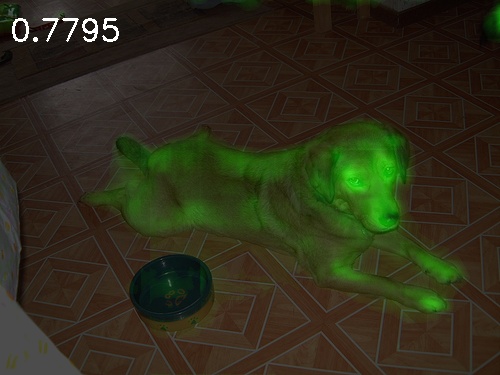}\includegraphics[width=2.8cm,height=2cm]{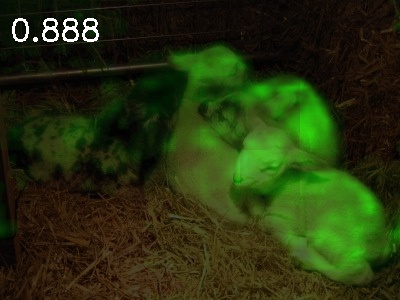}\includegraphics[width=2.8cm,height=2cm]{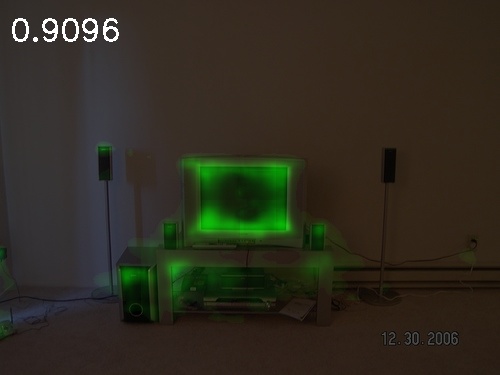} 
\par\end{centering}
\begin{centering}
\includegraphics[width=2.8cm,height=2cm]{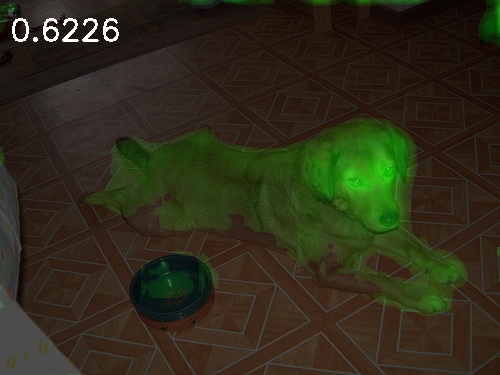}\includegraphics[width=2.8cm,height=2cm]{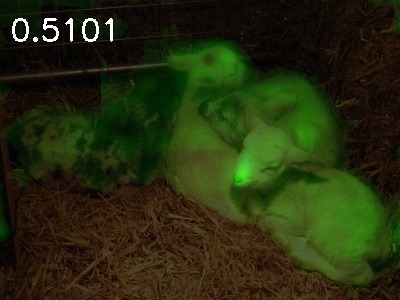}\includegraphics[width=2.8cm,height=2cm]{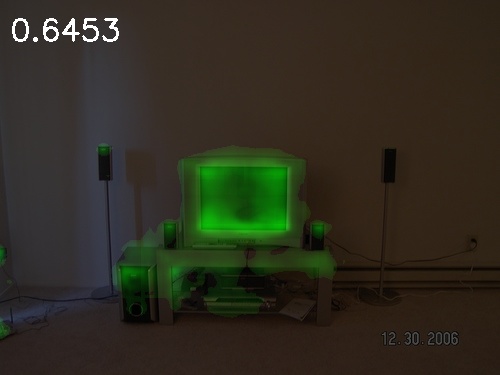} 
\par\end{centering}
\begin{centering}
\caption{Visualization of CAM and its Coefficient of Variation. Top: CAM. Bottom:
CAM after CVS. Only one category is highlighted in these images.}
\label{cvh} 
\par\end{centering}
\centering{} 
\end{figure}

\textbf{Improvements of Pseudo-mask Generation}: We evaluate the effectiveness
of CVS and PPMG in Tab. \ref{iucrf}. We firstly apply dense-CRF on
baseline (SEAM) and our variant in the top part. The mIoU of CAMs
increases by 1.52 and 0.98 respectively in these two settings. These
results are obtained from grid search of CRF background reduction
hyper parameter $\alpha$ from 0 to 20. In the middle part, we show
the individual gains and we find the gains of PPMG are more. In the
last line, PPMG with CVS and CRF improves the baseline by 4.6, and
the gain of our CAM is 3.28. In Fig. \ref{cvh} we give some examples
with its $c_{v}$ values and its results after CVS, we can observe that
the disparity problem is more critical with larger $c_{v}$ values
and CVS could handle it appropriately.

\begin{table}
\begin{centering}
\setlength\tabcolsep{15pt}
\begin{tabular}{ccc}
\hline 
Setting  & Baseline  & Ours\tabularnewline
\hline 
\hline 
CAM  & 52.72  & 58.21\tabularnewline
CRF  & 54.24  & 59.19\tabularnewline
\hline 
CRF \& CVS  & 54.41  & 60.83\tabularnewline
PPMG without CVS  & 57.02  & 61.23\tabularnewline
\hline 
PPMG  & \textbf{57.32}  & \textbf{61.49}\tabularnewline
\hline 
\end{tabular}
\par\end{centering}
\caption{\label{iucrf}CAM mIoU of different post-process settings on VOC12
val set. CVS: Coefficient of Variation Smoothing. PPMG: Proportional
Pseudo-mask Generation. Baseline CAM results are from SEAM, and ``Ours''
adds multi-crop and ScaleNet.}
\end{table}

\begin{figure*}[t]
\begin{centering}
\includegraphics[width=2cm,height=2.3cm]{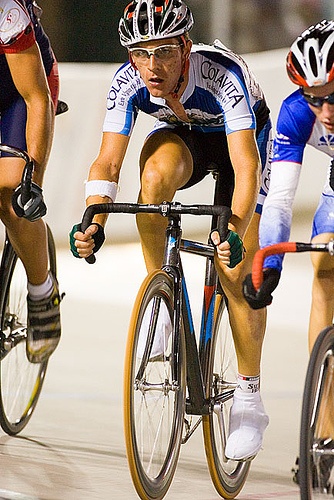}\includegraphics[width=3.5cm,height=2.3cm]{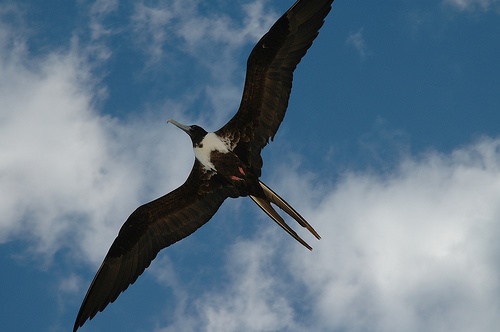}\includegraphics[width=2cm,height=2.3cm]{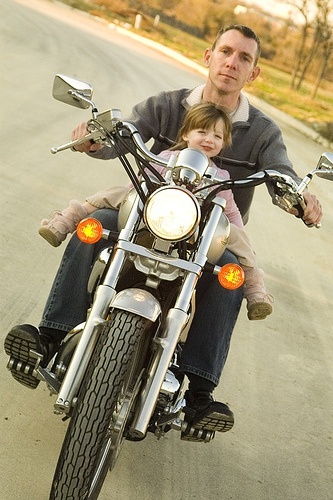}\includegraphics[width=3.5cm,height=2.3cm]{ori4}\includegraphics[width=3.5cm,height=2.3cm]{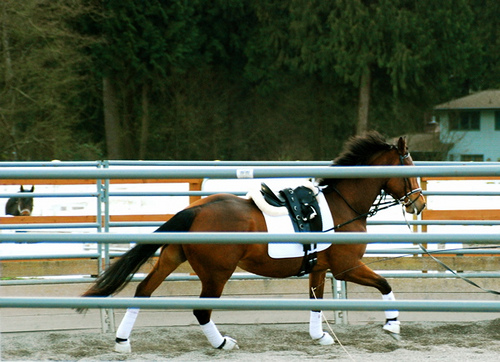}\includegraphics[width=2cm,height=2.3cm]{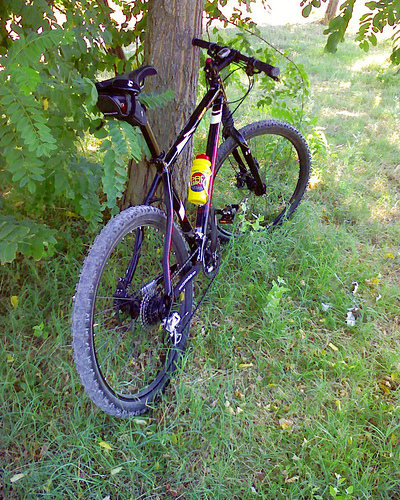} 
\par\end{centering}
\begin{centering}
\includegraphics[width=2cm,height=2.3cm]{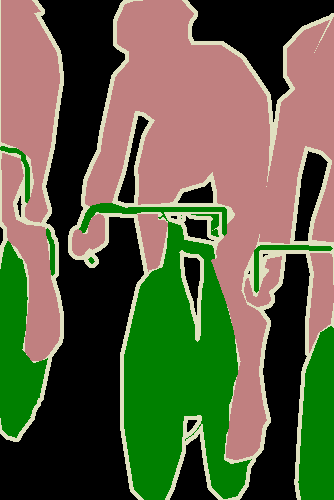}\includegraphics[width=3.5cm,height=2.3cm]{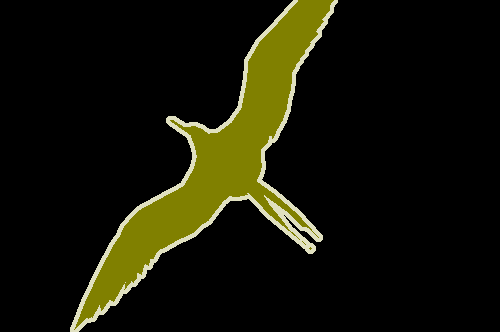}\includegraphics[width=2cm,height=2.3cm]{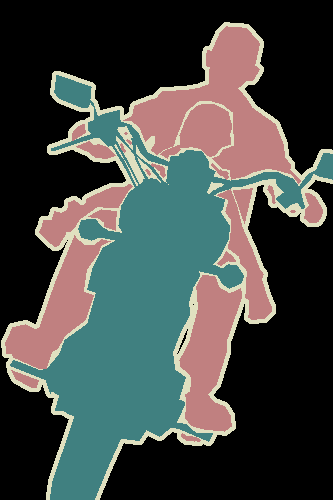}\includegraphics[width=3.5cm,height=2.3cm]{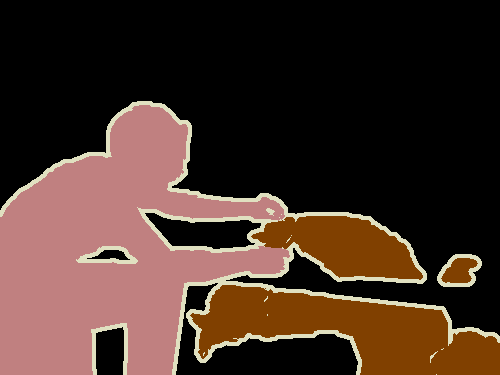}\includegraphics[width=3.5cm,height=2.3cm]{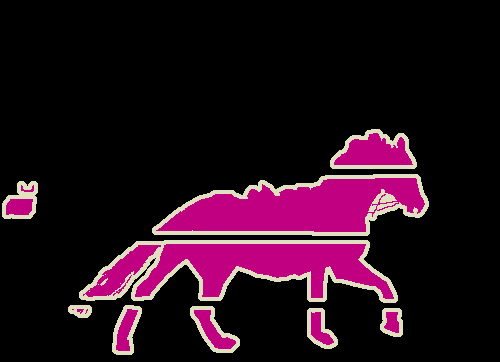}\includegraphics[width=2cm,height=2.3cm]{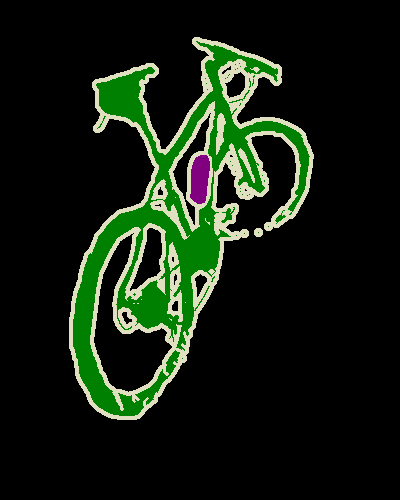} 
\par\end{centering}
\begin{centering}
\includegraphics[width=2cm,height=2.3cm]{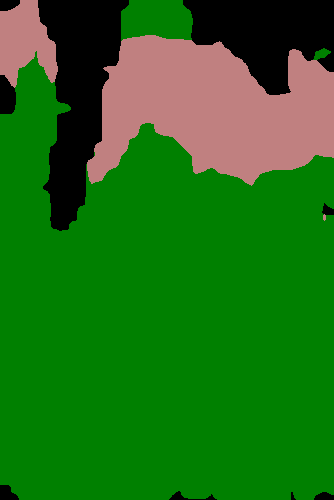}\includegraphics[width=3.5cm,height=2.3cm]{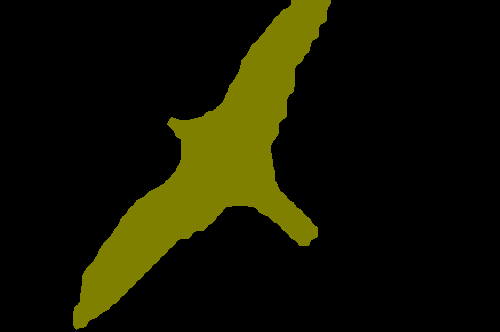}\includegraphics[width=2cm,height=2.3cm]{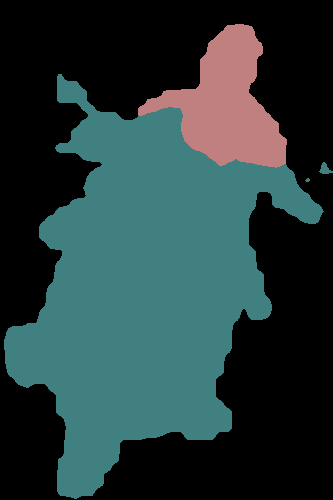}\includegraphics[width=3.5cm,height=2.3cm]{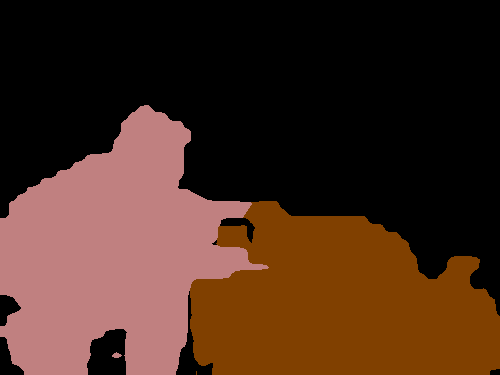}\includegraphics[width=3.5cm,height=2.3cm]{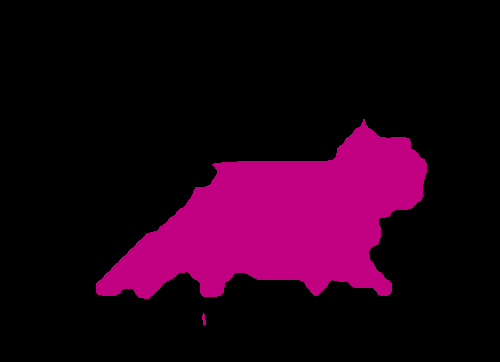}\includegraphics[width=2cm,height=2.3cm]{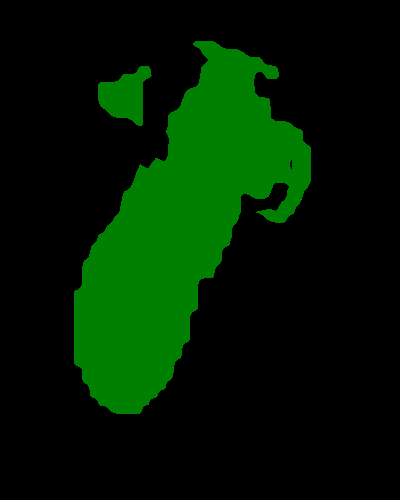} 
\par\end{centering}
\begin{centering}
\includegraphics[width=2cm,height=2.3cm]{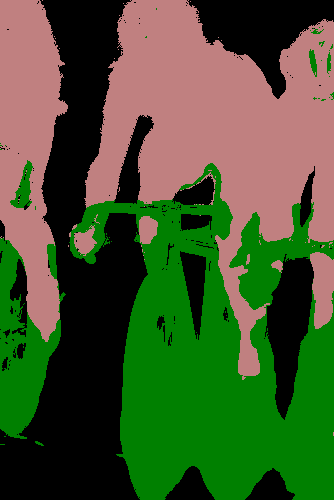}\includegraphics[width=3.5cm,height=2.3cm]{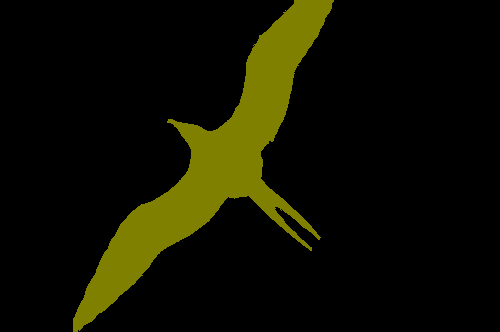}\includegraphics[width=2cm,height=2.3cm]{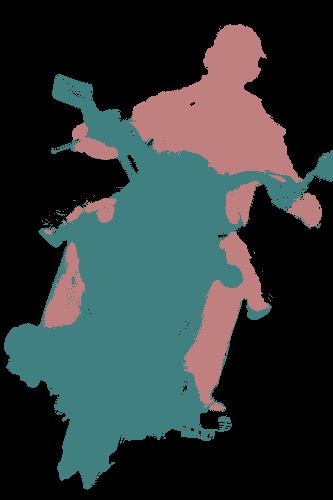}\includegraphics[width=3.5cm,height=2.3cm]{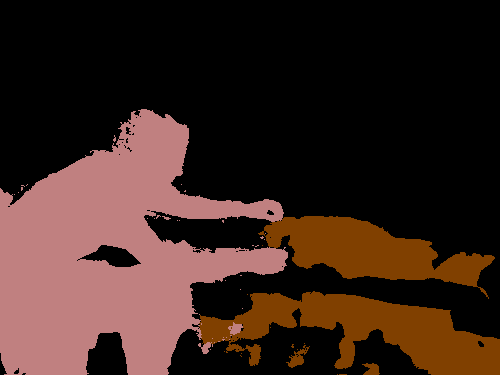}\includegraphics[width=3.5cm,height=2.3cm]{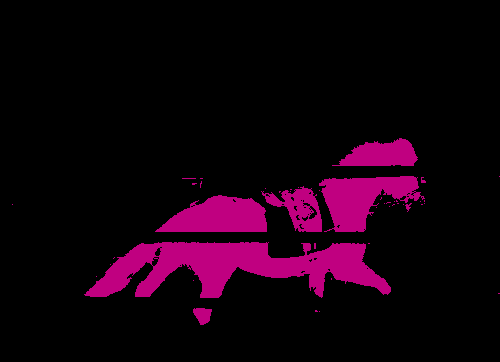}\includegraphics[width=2cm,height=2.3cm]{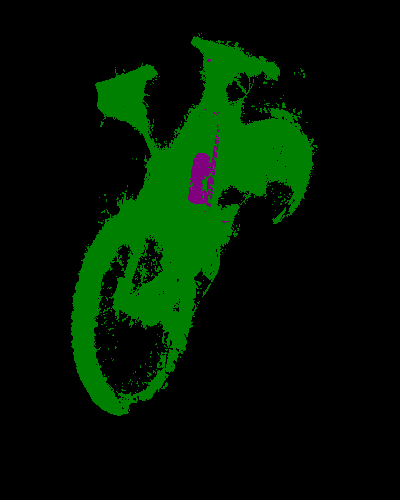} 
\par\end{centering}
\centering{}\caption{Qualitative results of pseudo-maskt. Top: original images in
VOC12 validation set. Second row: ground truth with ignored boundary.
Third row: pseudo-masks of SEAM (baseline) with Random Walk. Bottom
: our pseudo-masks from PPMG with CVS.}
\label{pseudo_vis} 
\end{figure*}

\textbf{Removal of Random Walk}: Most WSSS algorithms deploy Random
Walk to refine the CAM. It requires CRF operations at different $\alpha$
values to synthesize the training data. However,  
the results in Tab. \ref{refinement} suggest that the affinity network and Random Walk do not
work in our settings. So we remove the affinity network and Random
Walk in our pipeline. The visualization of our refined CAM is depicted
in Fig. \ref{pseudo_vis} with baseline SEAM.

\begin{table}
\begin{centering}\setlength\tabcolsep{7pt}
\begin{tabular}{ccccc}
\hline 
our CAM  & CRF  & RW  & PPMG  & VOC12 val\tabularnewline
\hline 
$\checked$  &  &  &  & 58.21\tabularnewline
$\checked$  & $\checked$  &  &  & 59.19\tabularnewline
$\checked$  &  & $\checked$  &  & 52.18\tabularnewline
$\checked$  &  &  & $\checked$  & \textbf{61.49}\tabularnewline
$\checked$  &  & $\checked$  & $\checked$  & 60.96\tabularnewline
\hline 
\end{tabular}
\par\end{centering}
\caption{\label{refinement}mIoU of different post-process settings.}
\end{table}

\textbf{Improvements of Pseudo-mask Utilization}: We firstly compare
three $pus$ operations in Tab. \ref{tab:ifl}. The top part shows
the results from different pseudo-masks without PUS. The middle part
shows the results of three $pus$ functions, among which, $pus_{clamp}$
performs best. Compared
to baseline pseudo-mask, the improvement rises from 0.45 to 3.46 after applying the
$pus_{clamp}$, which means the PUS is essential to our pipeline.

\begin{table}
\begin{centering}
\setlength\tabcolsep{15pt}
\begin{tabular}{ccc}
\hline 
$pus$  & Pseudo-Mask  & VOC12 val\tabularnewline
\hline 
\hline 
-  & SEAM \& RW  & 64.41\tabularnewline
-  & Ours  & 64.86\tabularnewline
\hline 
pow  & Ours  & 64.92\tabularnewline
ignore  & Ours  & 65.44\tabularnewline
clamp  & Ours  & \textbf{66.73}\tabularnewline
clamp  & SEAM \& RW  & 63.27\tabularnewline
\hline 
\end{tabular}
\par\end{centering}
\caption{\label{tab:ifl}Comparison of Pretended Under-fitting Strategy.}
\end{table}

We also validate the effectiveness of Cyclic Pseudo-mask in Tab. \ref{tab:cyclic}.
We find VOC dataset needs to update the pseudo-mask once while COCO
does not require cyclically updating.

\begin{table}
\begin{centering}
\setlength\tabcolsep{15pt}
\begin{tabular}{ccc}
\hline 
Cycle Times  & V0C val  & COCO val\tabularnewline
\hline 
0  & 66.73  & \textbf{36.73}\tabularnewline
1  & \textbf{68.50}  & 35.65\tabularnewline
2  & 68.27  & -\tabularnewline
\hline 
\end{tabular}
\par\end{centering}
\caption{\label{tab:cyclic}Performance of Cyclic Pseudo-mask in different
cycle times. 0 suggests non-cyclic pseudo-mask.}
\end{table}

In Tab. \ref{seg}, We show the accumulated gains of our methods,
including the classification, segmentation and pseudo-mask generation
methods. Although our classification pipeline improve the performance of CAM, application of RW instead of PPMG hurts the performance.  
\begin{center}
\begin{table}
\begin{centering}
\setlength\tabcolsep{5pt}
\begin{tabular}{cccccccc}
\hline 
Ori  & Cls  & RW  & PPMG  & PUS  & Cyclic  & R2N  & mIoU\tabularnewline
\hline 
$\checked$  &  &  &  &  &  &  & 52.72\tabularnewline
$\checked$  & $\checked$  &  &  &  &  &  & 58.21\tabularnewline
$\checked$  & $\checked$  & $\checked$  &  &  &  &  & 52.18\tabularnewline
$\checked$  & $\checked$  &  & $\checked$  &  &  &  & 61.49\tabularnewline
$\checked$  & $\checked$  &  & $\checked$  & $\checked$  &  &  & 66.73\tabularnewline
$\checked$  & $\checked$  &  & $\checked$  & $\checked$  & $\checked$  &  & 68.50\tabularnewline
$\checked$  & $\checked$  &  & $\checked$  & $\checked$  & $\checked$  & $\checked$  & 70.01\tabularnewline
\hline 
\end{tabular}
\par\end{centering}
\caption{\label{seg}Accumulated gains on V0C12 validation set. Cls means our
classification setting. RW is affinity with Random Walk. PPMG is Proportional
Pesudo-mask Generation with CVS. PUS is fully-supervised segmentation
with Pretended Under-fitting Strategy. Cyclic indicates Cyclic Pseudo-mask.
R2N change the backbone of segmentation to Res2Net.}
\end{table}
\par\end{center}

\subsection{Comparison with State-of-the-arts}

We compare the final results after fully supervised segmentation step
in Tab. \ref{comp1}. The backbones of segmentation framework are
listed in the table. Our method does not require affinity network and
Random Walk inference, while achieves new state-of-the-arts in both
VOC12 and COCO14 validation datasets at 70.0 and 40.2 respectively.
Compared to our baseline SEAM, the gain is 5.5 in VOC12 and 5.0 in
COCO14. For the same backbone, the improvements are 4.0 and 5.0 respectively. Note that the performance of ScaleNet101 on VOC is lower than base model ResNet38, but it suits tiny objects well on COCO at mIoU 40.2.

\begin{table}
\begin{centering}
\setlength\tabcolsep{4pt}
\begin{tabular}{ccc|ccc}
\hline 
Method & Backbone & RW & val & test & COCO\tabularnewline
\hline 
\hline 
BFBP \cite{saleh2016built} & VGG16 & $\mathsf{x}$ & 46.6 & 48.0 & 20.4\tabularnewline
SEC \cite{kolesnikov2016seed} & VGG16 & $\mathsf{x}$ & 50.7 & 51.7 & 22.4\tabularnewline
AffinityNet \cite{ahn2018learning} & ResNet38 & $\checked$ & 61.7 & 63.7 & -\tabularnewline
IRNet \cite{8953768} & ResNet50 & $\checked$ & 63.5 & 64.8 & -\tabularnewline
OAA \cite{jiang2019integral} & ResNet101 & $\mathsf{x}$ & 63.9 & 65.6 & -\tabularnewline
ICD \cite{fan2020learning} & ResNet101 & $\mathsf{x}$ & 64.1 & 64.3 & -\tabularnewline
{SEAM \cite{Wang_2020_CVPR}}$^{\star}$ & ResNet38 & $\checked$ & 64.4 & 65.7 & -\tabularnewline
\textcolor{blue}{SEAM \cite{Wang_2020_CVPR}} & ResNet38 & $\checked$ & \textcolor{blue}{64.5} & \textcolor{blue}{65.7} & \textcolor{blue}{31.7}\tabularnewline
SSDD \cite{shimoda2019self} & ResNet & $\checked$ & 64.9 & 65.5 & -\tabularnewline
CONTA \cite{zhang2020causal} & ResNet38 & $\checked$ & 66.1 & 66.7 & 32.8\tabularnewline
SC-CAM \cite{chang2020weakly} & ResNet101 & $\checked$ & 66.1 & 65.9 & -\tabularnewline
Sun \etal \cite{sun2020mining} & ResNet101 & $\checked$ & 66.2 & 66.9 & -\tabularnewline
\hline 
PMM & ResNet38 & $\mathsf{x}$ & 68.5 & 69.0 & 36.7\tabularnewline
PMM & ScaleNet101 & $\mathsf{x}$ & 67.1 & 67.7 &
\textbf{40.2}\tabularnewline
PMM & Res2Net101 & $\mathsf{x}$ & \textbf{70.0} & \textbf{70.5} &
35.7\tabularnewline
\hline 
\end{tabular}
\par\end{centering}
\caption{\label{comp1}Performance comparisons with state-of-the-art WSSS methods
on VOC 2012 and COCO 2014 validation datasets. All the methods listed
use image-level supervision only without extra models. Checkmarks
suggest these methods need training an affinity network and Random
Walk inference. SEAM in blue is our baseline and$^{\star}$ indicates
our re-implementation baseline in same segmentation code.}
\end{table}

In Tab. \ref{comp2} we compare our PMM to methods with extra
information like saliency model, bounding box supervision, extra dataset
and segment-based object proposals. These methods introduce more information
and thus take advantage to the methods in Tab. \ref{comp1}. Even
though, our PMM achieves new SOTA in both VOC12 and COCO14 without extra
information. Especially, on COCO14 our method significantly surpasses
the best published results by 3.1.

\begin{table}
\begin{centering}
\setlength\tabcolsep{6pt}
\begin{tabular}{cc|ccc}
\hline 
Method & Extra Info & val & test & COCO\tabularnewline
\hline 
\hline 
DSRG \cite{huang2018weakly} & MSRA-B & 61.4 & 63.2 & 26.0$^{\dag}$\tabularnewline
BoxSup \cite{dai2015boxsup} & $\mathcal{D}$ & 62.0$^{\dag}$ & 64.6 & -\tabularnewline
FickleNet \cite{lee2019ficklenet} & $\mathcal{S}$ & 64.9 & 65.3 & -\tabularnewline
SDI \cite{Khoreva_2017_CVPR} & $\mathcal{D}$+BSDS & 65.7 & 67.5 & -\tabularnewline
OAA$^{+}$ \cite{jiang2019integral} & $\mathcal{S}$ & 65.2 & 66.4 & \tabularnewline
SGAN \cite{yao2020saliency} & $\mathcal{S}$ & 67.1 & 67.2 & 33.6\tabularnewline
ICD \cite{fan2020learning} & $\mathcal{S}$ & 67.8 & 68.0 & -\tabularnewline
Li \etal \cite{li2020group} & $\mathcal{S}$ & 68.2 & 68.5 & 28.4$^{\dag}$\tabularnewline
LIID \cite{liu2020leveraging} & SOP & 66.5 & 67.5 & -\tabularnewline
LIID \cite{liu2020leveraging} & SOP & 69.4$^{\ddagger}$ & 70.4 & -\tabularnewline
\hline 
PMM & - & 68.5 & 69.0 & 36.7\tabularnewline
PMM & - & 67.1$^{\star}$ & 67.7$^{\star}$ & \textbf{40.2$^{\star}$}\tabularnewline
PMM & - & \textbf{70.0}$^{\ddagger}$ & \textbf{70.5}$^{\ddagger}$ & 35.7$^{\ddagger}$\tabularnewline
\hline 
\end{tabular}
\par\end{centering}
\centering{}\caption{\label{comp2}Performance comparisons state-of-the-art WSSS methods
on VOC 2012 and COCO 2014 validation datasets. Except PMM, other methods
use different supervision or extra models. $\mathcal{S}$ means external
saliency models and $\mathcal{D}$ means supervision of detection.
SOP is segment-based object proposals. $\dag$ indicates using VGG, $\ddagger$ indicates Res2Net and $^{\star}$ means ScaleNet others using ResNet.}
\end{table}

\begin{figure*}[t]
\begin{centering}
\includegraphics[width=2.8cm,height=2cm]{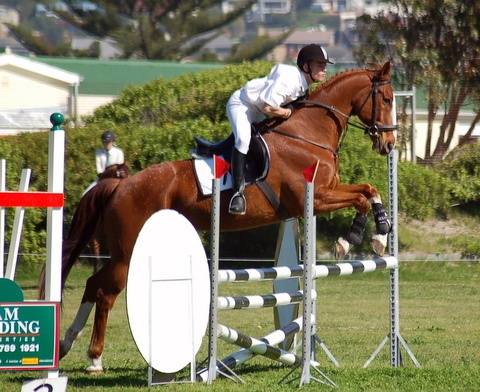}\includegraphics[width=2.8cm,height=2cm]{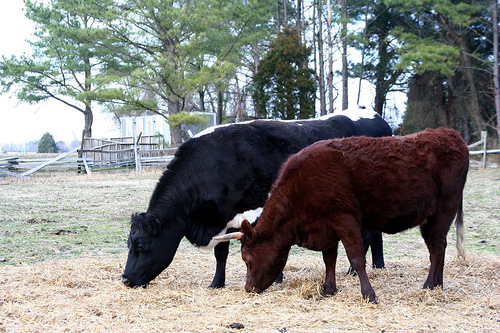}\includegraphics[width=2.8cm,height=2cm]{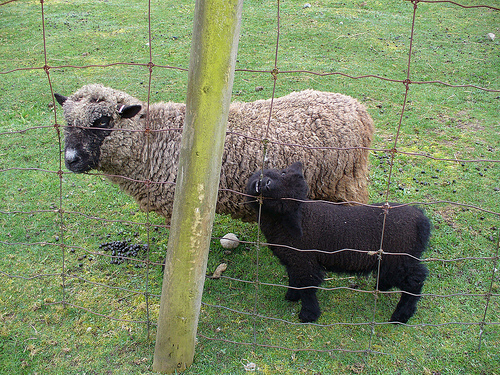}\includegraphics[width=2.8cm,height=2cm]{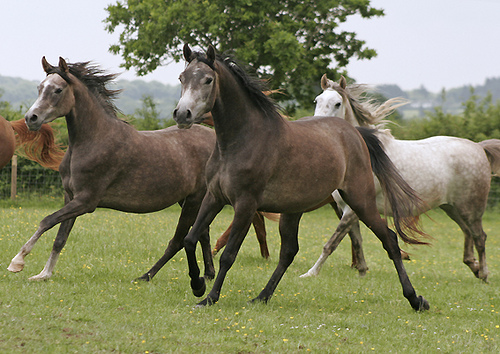}\includegraphics[width=2.8cm,height=2cm]{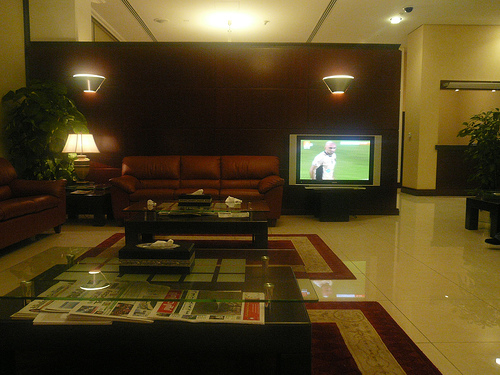}\includegraphics[width=2.8cm,height=2cm]{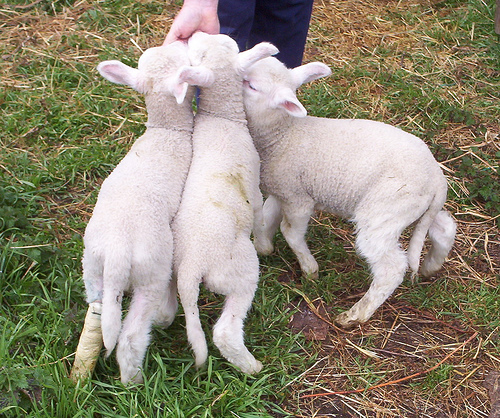} 
\par\end{centering}
\begin{centering}
\includegraphics[width=2.8cm,height=2cm]{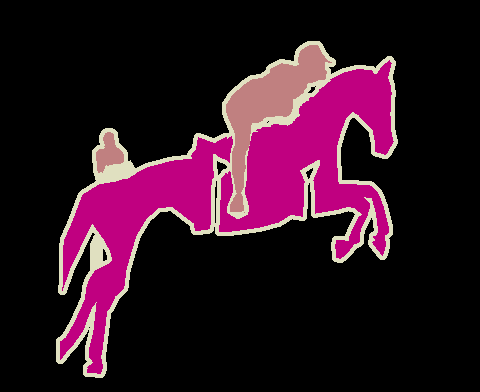}\includegraphics[width=2.8cm,height=2cm]{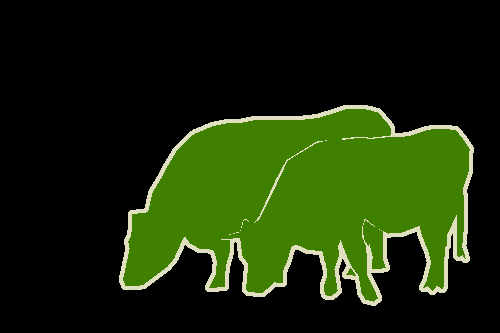}\includegraphics[width=2.8cm,height=2cm]{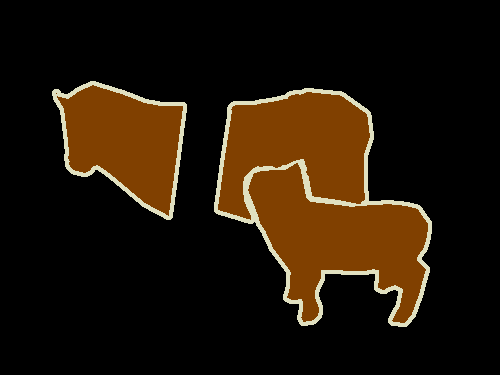}\includegraphics[width=2.8cm,height=2cm]{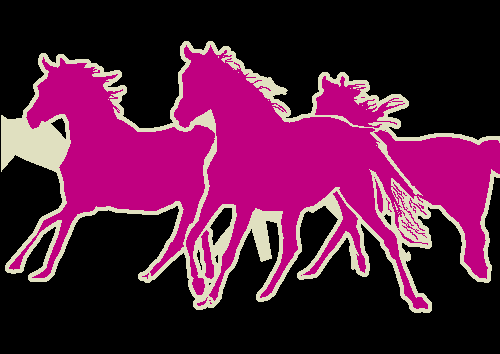}\includegraphics[width=2.8cm,height=2cm]{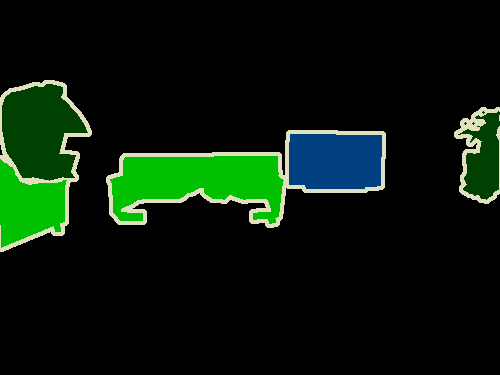}\includegraphics[width=2.8cm,height=2cm]{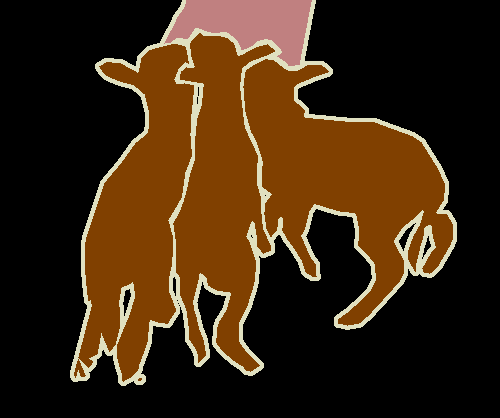} 
\par\end{centering}
\begin{centering}
\includegraphics[width=2.8cm,height=2cm]{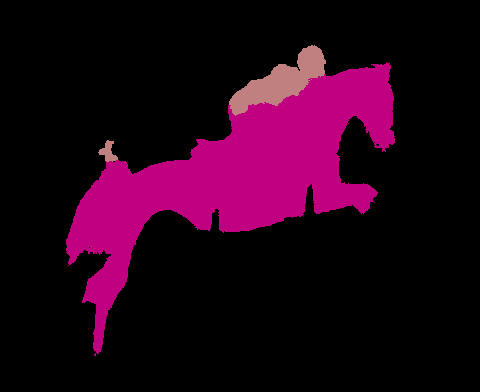}\includegraphics[width=2.8cm,height=2cm]{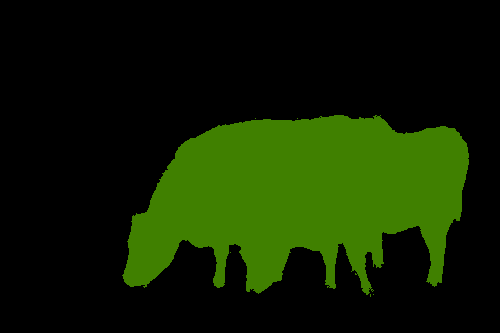}\includegraphics[width=2.8cm,height=2cm]{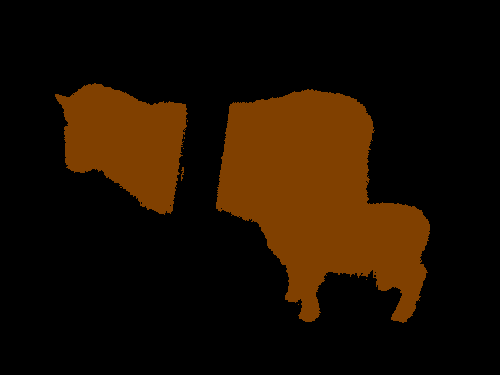}\includegraphics[width=2.8cm,height=2cm]{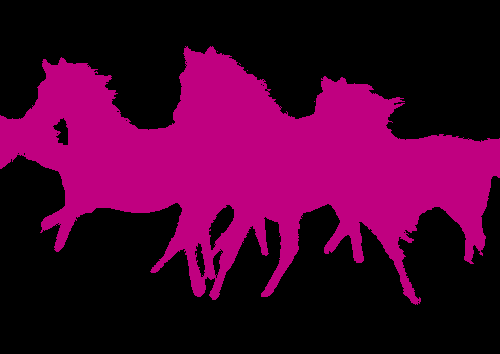}\includegraphics[width=2.8cm,height=2cm]{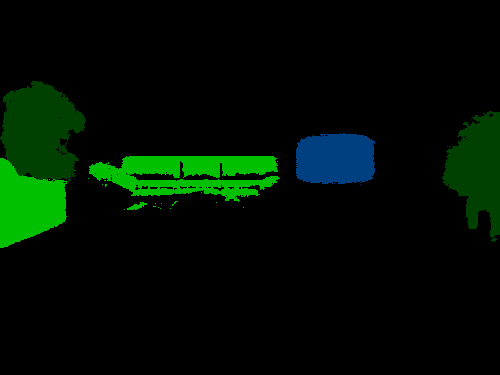}\includegraphics[width=2.8cm,height=2cm]{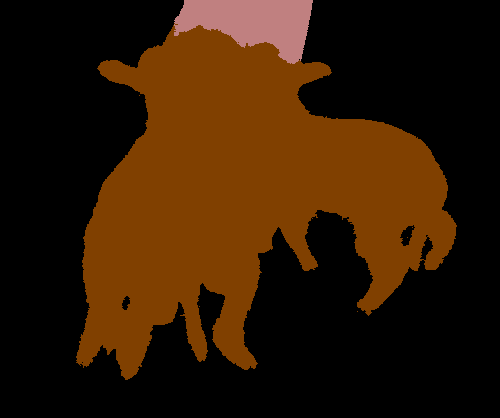} 
\par\end{centering}
\begin{centering}
\caption{Qualitative segmentation results on the VOC12 validation set. Top:
original images. Middle: ground truth with ignored boundary. Bottom:
Our results of PMM.}
\label{vis_seg} 
\par\end{centering}
\centering{} 
\end{figure*}

In Fig. \ref{vis_seg}, we shows some qualitative results on VOC12
validation set, which verifies the effectiveness of our PMM.

\subsection{Failure Case and Remedy}

\begin{figure}
\begin{centering}
\includegraphics[width=2.7cm,height=1.5cm]{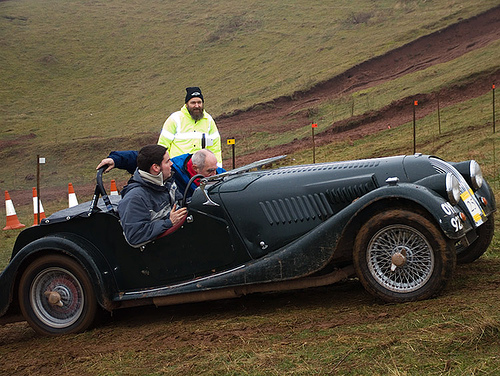}\includegraphics[width=2.7cm,height=1.5cm]{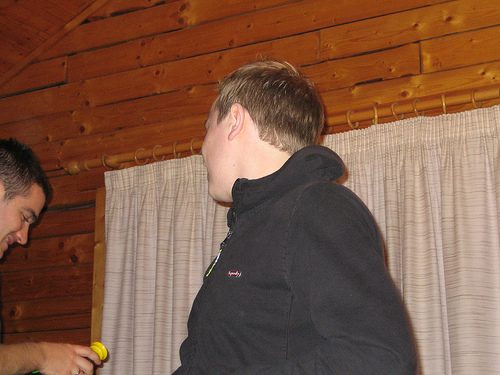}\includegraphics[width=2.7cm,height=1.5cm]{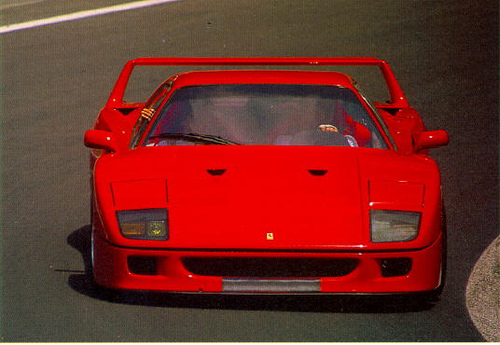} 
\par\end{centering}
\begin{centering}
\includegraphics[width=2.7cm,height=1.5cm]{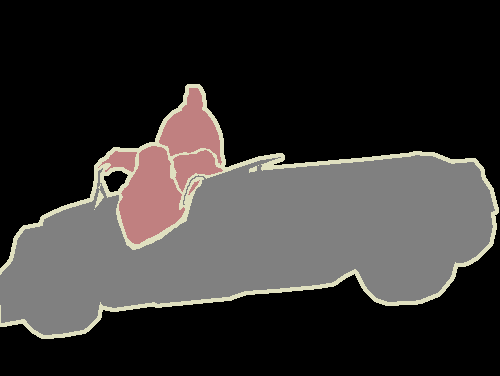}\includegraphics[width=2.7cm,height=1.5cm]{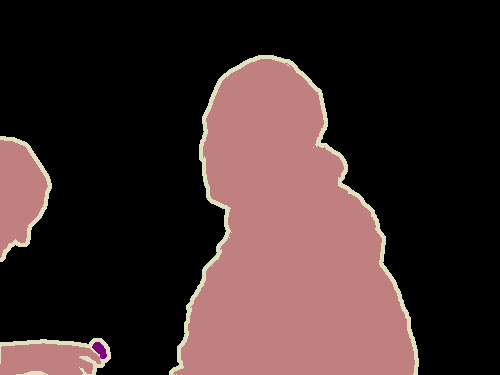}\includegraphics[width=2.7cm,height=1.5cm]{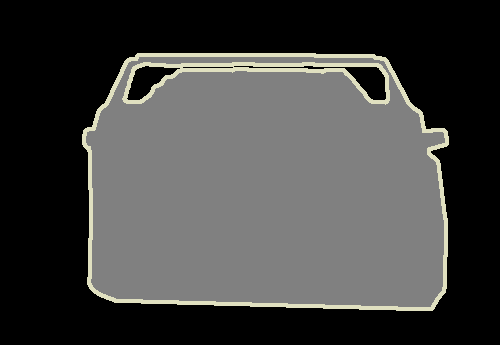} 
\par\end{centering}
\begin{centering}
\includegraphics[width=2.7cm,height=1.5cm]{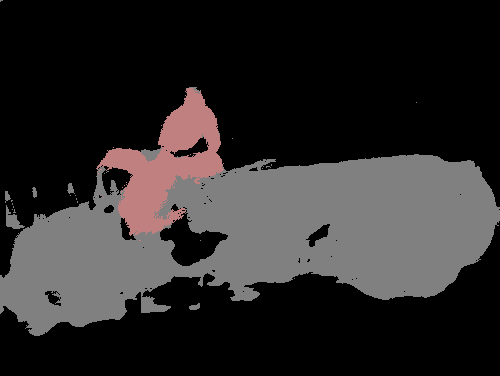}\includegraphics[width=2.7cm,height=1.5cm]{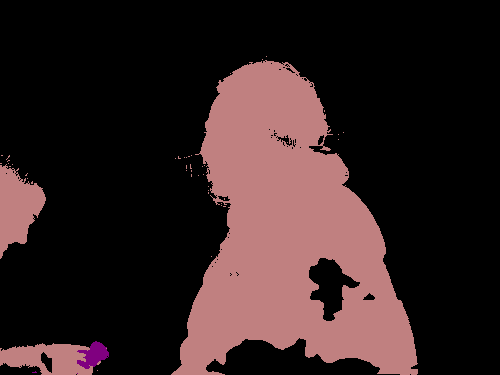}\includegraphics[width=2.7cm,height=1.5cm]{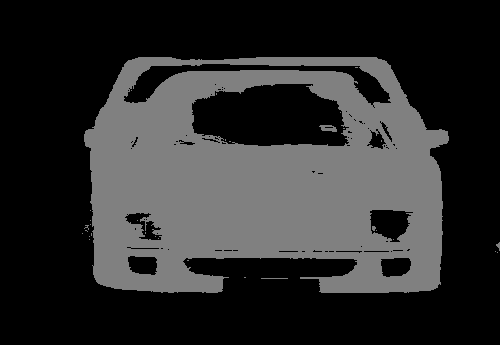} 
\par\end{centering}
\begin{centering}
\includegraphics[width=2.7cm,height=1.5cm]{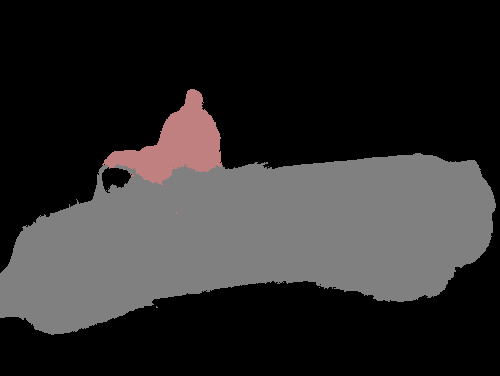}\includegraphics[width=2.7cm,height=1.5cm]{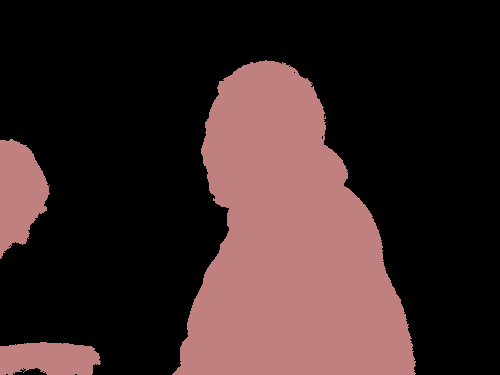}\includegraphics[width=2.7cm,height=1.5cm]{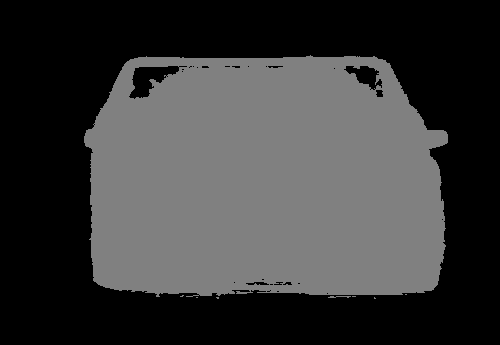} 
\par\end{centering}
\caption{Failure cases of pseudo-masks and its segmentation results on VOC12
validation set. First line: original images. Second line: ground truth.
Third line: failure cases of our pseudo-masks. Last line: segmentation
remedy results.}

\label{remedy} 
\end{figure}

In Fig. \ref{pseudo_vis} we can see that, the details of our pseudo-masks
are much better than SEAM with Random Walk. But the mIoU of ours is
not very high. The problem is the under-activation CAM, which causes
incomplete prediction map or false positive. For this issue, the experiment
in Tab. \ref{seg} and visualization in Fig. \ref{remedy} prove that
this phenomenon is able to remedy by fully supervised segmentation. Thus PPMG reserves the details, on the other hand, segmentation model remedies the miss area.

\section{Conclusion}

In weakly supervised semantic segmentation, there are some matters
about pseudo-mask in the generation and utilization. To solve these
matters, we propose the Proportional Pseudo-mask Generation to identify
the category independently and avoid direct score comparison, and the
Coefficient of Variation Smoothing to smooth the CAM by its distribution
statistics. Then, we add Pretended Under-fitting Strategy to FSSS,
and verify the effectiveness of Cyclic Pseudo-mask experimentally.
We solve the matters about pseudo-mask and achieves new state-of-the-art
in both VOC12 and COCO14 datasets, even beyond the methods using extra
information.

\section{Acknowledgement}

The work is partially supported by Innovation and Technology Commission of the Hong Kong Special Administrative Region, China (Enterprise Support Scheme under the Innovation and Technology Fund B/E030/18).

{\small
\bibliographystyle{ieee_fullname}
\bibliography{egbib}
}

\end{document}